\documentclass[useAMS,usenatbib]{coin}
\usepackage{algorithm}
\usepackage{algorithmic}
\usepackage{multirow,rotating}
\usepackage{amsmath}
\usepackage{graphicx,color}

\newenvironment{singlespacedlist}{
\begin{itemize}{}{}
	\setlength{\itemsep}{4pt}
	\setlength{\parskip}{0pt}
	\setlength{\parsep}{0pt}}{\end{itemize}
}

\newcommand{\true}{{\bf true}}
\newcommand{\false}{{\bf false}}
\newcommand{\inserthalfnewline}{\vspace{-2mm} \\} 

\title[Missing Value Imputation With Unsupervised Backpropagation]{Missing Value Imputation With Unsupervised Backpropagation}

\author{{\sc Michael~S.~Gashler}\\
		{mgashler@uark.edu}\\
        {\it Department of Computer Science and Computer Engineering,}\\
        {\it University of Arkansas, Fayetteville, Arkansas, USA.}\\ 
		\and
        {\sc Michael~R.~Smith, Richard~Morris, Tony~Martinez}\\
        {msmith@axon.cs.byu.edu, rmorris@axon.cs.byu.edu, martinez@cs.byu.edu}\\
        {\it Department of Computer Science,}\\
        {\it Brigham Young University, Provo, Utah, USA.}}

\begin{document}

\pagerange{\pageref{firstpage}--\pageref{lastpage}} \pubyear{2013}

\volume{64}
\artmonth{July}
\doi{10.1111/j.0000-0000.2013.00000.x}


\label{firstpage}


\begin{abstract}
Many data mining and data analysis techniques operate on dense matrices or complete tables of data.
Real-world data sets, however, often contain unknown values.
Even many classification algorithms that are designed to operate with missing values still exhibit deteriorated accuracy.
One approach to handling missing values is to fill in (impute) the missing values.
In this paper, we present a technique for unsupervised learning called \emph{Unsupervised Backpropagation} (UBP), which
trains a multi-layer perceptron to fit to the manifold sampled by a set of observed point-vectors.
We evaluate UBP with the task of imputing missing values in datasets, and show that UBP is able to
predict missing values with significantly lower sum-squared error than other collaborative filtering and imputation techniques.
We also demonstrate with 24 datasets and 9 supervised learning algorithms that classification accuracy is
usually higher when randomly-withheld values are imputed using UBP, rather than with other methods.
\end{abstract}

%
%


\maketitle
\vspace*{-12pt}
\section{Introduction}
\label{s:intro}

Many effective machine learning techniques are designed to operate on dense matrices or complete tables of data.
Unfortunately, real-world datasets often include only samples of observed values mixed with many missing or unknown elements.
Missing values may occur due to human impatience, human error during data entry, data loss, faulty sensory equipment, changes in data collection methods, inability to decipher handwriting, privacy issues, legal requirements, and a variety of other practical factors.
Thus, improvements to methods for imputing missing values can have far-reaching impact on improving the effectiveness of existing learning algorithms for operating on real-world data.
We present a method for imputation called \emph{Unsupervised Backpropagation} (UBP), which trains a multi-layer perceptron (MLP) to fit to the manifold represented by the known features in a dataset.
We demonstrate this algorithm with the task of imputing missing values, and we show that it is significantly more effective than other methods for imputation.

Backpropagation has long been a popular method for training neural networks \citep{rumelhart:backpropagation,werbos:backpropagation}.
A typical supervised approach trains the weights, $\mathbf{W}$, of a multilayer perceptron (MLP) to fit to a set of training examples, consisting of a set of $n$ feature vectors $\mathbf{X}=\langle \mathbf{x}_1, \mathbf{x}_2, ..., \mathbf{x}_n \rangle$, and $n$ corresponding label vectors $\mathbf{Y}=\langle \mathbf{y}_1, \mathbf{y}_2, ..., \mathbf{y}_n \rangle$.
With many interesting problems, however, training data is not available in this form.
In this paper, we consider the significantly different problem of training an MLP to estimate, or impute, the missing attribute values of $\mathbf{X}$ .
Here $\mathbf{X}$ is represented as an $n \times d$ matrix where each of the $d$ attributes may be continuous or categorical.
Because the missing elements in $\mathbf{X}$ must be predicted, $\mathbf{X}$ becomes the output of the MLP, rather than the input.
A new set of latent vectors, $\mathbf{V}=\langle \mathbf{v}_1, \mathbf{v}_2, ..., \mathbf{v}_n \rangle$, will be fed as inputs into the MLP.
However, no examples from $\mathbf{V}$ are given in the training data.
Thus, both $\mathbf{V}$ and $\mathbf{W}$ must be trained using only the known elements in $\mathbf{X}$.
After training, each $\mathbf{v}_i$ may be fed into the MLP to predict all of the elements in $\mathbf{x}_i$.

Training in this manner causes the MLP to fit a surface to the (typically non-linear) manifold sampled by $\mathbf{X}$.
After training, $\mathbf{V}$ may be considered as a reduced-dimensional representation of $\mathbf{X}$.
That is, $\mathbf{V}$ will be an $n \times t$ matrix, where $t$ is typically much smaller than $d$, and the MLP maps $\mathbf{V}\mapsto\mathbf{X}$.

UBP accomplishes the task of training an MLP using only the known attribute values in $\mathbf{X}$ with on-line backpropagation.
For each presentation of a known value of the $c^\text{th}$ attribute
from the $r^\text{th}$ instance ($x_{r,c}\in\mathbf{X}$), UBP simultaneously computes a
gradient vector $\mathbf{g}$ to update the weights $\mathbf{W}$, and a gradient
vector $\mathbf{h}$ to update the input vector $\mathbf{v}_r$.
($x_{r,c}$ is the element
in row $r$, column $c$ of $\mathbf{X}$.)

In this paper, we demonstrate UBP as a method for imputing missing values, and show that it outperforms other approaches at this task.
We compare UBP against 5 other imputation methods on a set of 24 data sets.
10\% to 90\% of the values are removed from the data sets completely at random.
We show that UBP predicts the missing values with signficantly lower error (as measured by sum-squared difference with normalized values) than other approaches.
We also evaluated 9 learning algorithms to compare classification accuracy using imputed data sets.
Learning algorithms using imputed data from UBP usually achieve higher classification accuracy than with any of the other methods.
The increase is most significant when 30\% to 70\% of the data is missing.

The remainder of this paper is organized as follows.
Section \ref{sec_related} reviews related work to UBP and missing value imputation.
UBP is described in Section \ref{sec_UBP}.
Section \ref{sec_results} presents the results of comparing UBP with other imputation methods.
We provide conclusions and a discussion of future directions for UBP in Section \ref{sec_conclusions}.

\section{Related Work}\label{sec_related}

As an algorithm, UBP falls at the intersection of several different paradigms: neural networks, collaborative filtering, data imputation, and manifold learning.
In neural networks, UBP is an extension of generative backpropagation \citep{hinton:generative_backpropagation}.
Generative backpropagation adjusts the inputs in a neural network while holding the weights constant.
UBP, by contrast, computes both the weights and the input values simultaneously.
Related approaches have been used to generate labels for images \citep{coheh:label_images_generative_backprop}, and for natural language \citep{bengio:neural_language_generative_backprop}.
Although these techniques have been used for labeling images and documents, to our knowledge, they have not been used for the application of imputing missing values.
UBP differs from generative backpropagation in that it trains the weights simultaneously with the inputs, instead of training them as a pre-processing step.

UBP may also be classified as a manifold learning algorithm.
Like common non-linear dimensionality reduction (NLDR) algorithms, such as Isomap \citep{tenenbaum:isomap}, MLLE \citep{zhang:mlle}, or Manifold Sculpting \citep{gashler:manifoldsculpting}, it reduces a set of high-dimensional vectors, $\mathbf{X}$, to a corresponding set of low-dimensional vectors, $\mathbf{V}$.
Unlike these algorithms, however, UBP also learns a model of the manifold.
Also unlike these algorithms, UBP is designed to operate with incomplete observations.

In collaborative filtering, UBP may be viewed as a non-linear generalization of matrix factorization (MF).
MF is a linear dimensionality reduction technique that can be effective for collaborative filtering \citep{adomavicius:recommender_systems} as well as imputation.
This method has become a popular technique, in part due to its effectiveness with the data used in the NetFlix competition \citep{koren:matrix_factorization}.
MF involves factoring the data matrix into two much-smaller matrices.
These smaller matrices can then be combined to predict all of the missing values in the original dataset.
It is equivalent to using linear regression to project the data onto its first few principal components.
Unfortunately, MF is not well-suited for data that exhibits non-linearities.
It was previously shown that matrix factorization could be represented with a neural network model involving one hidden layer and linear activation functions \citep{takacs2009scalable}.
In comparison with this approach, UBP uses a standard MLP with an arbitrary number of hidden layers and non-linear activation functions, instead of the network structure previously proposed for matrix factorization.
MF produces very good results at the task of imputation, but we demonstrate that UBP does better.

As an imputation technique, UBP is a refinement of Nonlinear PCA \citep{nlpca} (NLPCA), which has been shown to be effective for imputation.
This approach also uses gradient descent to train an MLP to map from low to high-dimensional space.
After training, the weights of the MLP can be used to represent non-linear components within the data.
If these components are extracted one-at-a-time from the data, then they are the principal components, and NLPCA becomes a non-linear generalization of PCA.
Typically, however, these components are all learned together, so it would more properly be termed a non-linear generalization of MF.
NLPCA was evaluated with the task of missing value imputation \citep{nlpca}, but its relationship to MF was not yet recognized at the time, so it was not compared against MF.
One of the contributions of this paper is that we show NLPCA to be a significant improvement over MF at the task of imputation.
We also demonstrate that UBP achieves even better results than NLPCA at the same task, and is the best algorithm for imputation of which we are aware.
The primary difference between NLPCA and UBP is that UBP utilizes a three-phase training approach (described in Section~\ref{sec_UBP}) which makes it more robust against falling into a local optimum during training.

UBP is comparable with the latter-half of an autoencoder \citep{autoencoders}. Autoencoders
create a low dimensional representation of a training set by using the
training examples as input features as well as the target values. The
first half of the encoder reduces the input features into low
dimensional space by only have $n$ nodes in the middle layer where $n$
is less than the number of input features. The latter-half of the
autoencoder then maps the low dimensional representation of the training
set back to the original input features. However, to capture non-linear
dependencies in the data, autoencoders require deep architectures to
allow for layers between the inputs and low dimensional representation
of the data and between the low dimensional representation of the data
and the output. The deep architecture makes training an autoencoder
difficult and computationally expensive, generally requiring
unsupervised layer-wise training \citep{Bengio2007, Erhan2009}. Because
UBP trains a network with half the depth of a corresponding autoencoder,
UBP is practical for many problems for which autoencoders are too
computationally expensive.

Since we demonstrate UBP with the application of imputing missing values in data, it is also relevant to consider other approaches that are classically used for this task.
Simple methods, such as dropping patterns that contain missing values or randomly drawing values to replace the missing values, are often used based on simplicity for implementation.
These methods, however, have significant obvious disadvantages when data is scarce.
Another common approach is to treat missing elements as having a unique value.
This approach, however has been shown to bias the parameter estimates for multiple linear regression models \citep{Jones1996} and to cause problems for inference with many models \citep{Shafer1997}.
We take it for granted that better accuracy is desirable, so these methods should generally not be used, as better methods do exist.

A simple improvement over BL is to compute a separate centroid for each output class.
The disadvantages of this method are that it is not suitable for regression problems, and it cannot generalize to unlabeled data since it depends on labels to impute.
Methods based on maximum likelihood \citep{little2002statistical} have long been studied in statistics, but these also depend on pattern labels.
Since it is common to have more unlabeled data than labeled data, we restrict our analysis to unsupervised methods that do not rely on labels to impute missing values.

Another well-studied approach involves training a supervised learning algorithm to predict missing values using the non-missing values as inputs \citep{quinlain_missing_values,lakshminarayan1996imputation,Farhangfar2008}.
Unfortunately, the case where multiple values are missing in one pattern present a difficulty for these approaches.
Either a learning algorithm must be used that implicitly handles missing values in some manner, or an exponential number of models must be trained to handle each combination of missing values.
Further, it has also been shown that results with these methods tend to be poor when there are high percentages (more than about 15\%) of missing values \citep{Acuna2004}.

One very effective collaborative filtering method for imputation is to cluster the data, and then make predictions according to the centroid of the cluster in which each point falls \citep{adomavicius:recommender_systems}.
Luengo compared several imputation methods by evaluating their effect on classification accuracy \citep{luengo:thesis2011}.
He found cluster-based imputation with Fuzzy $k$-Means (FKM) \citep{li_fuzzy_k_means} using Manhattan distance to outperform other methods, including those involving state of the art machine learning methods and other methods traditionally used for imputation.
Our analysis, however, finds that most of the methods we compared outperform FKM.

A related imputation method called instance-based imputation (IBI) is to combine the non-missing values of the $k$-nearest neighbors of a point to replace its missing values.
To evaluate the similarity between points, cosine correlation is often used because it tends to be effective in the presence of missing values \citep{adomavicius:recommender_systems,li:filtering_techniques,sarwar:item_based_collab_filter}.

UBP, as well as the aforementioned imputation techniques, are considered single imputation techniques because only one imputation for each missing value is made.
Single imputation has the disadvantage of introducing large amounts of bias since the imputed values do not reflect the added uncertainty from the fact that values are missing.
To overcome this, \citet{rubin:1987} proposed multiple imputation that estimates the added variance by combining the outcomes of $I$ imputed data sets.
Similarly, ensemble techniques have also been shown to be effective for imputing missing values \citep{schafer2002missing}.
In this paper, we do not compare against ensemble methods because UBP involves a single model, and it may be included in an ensemble as well as any other imputation method.



\section{Unsupervised Backpropagation}
\label{sec_UBP}

\begin{figure*}[tbh]
	\begin{center}
		\includegraphics[width=4.6in]{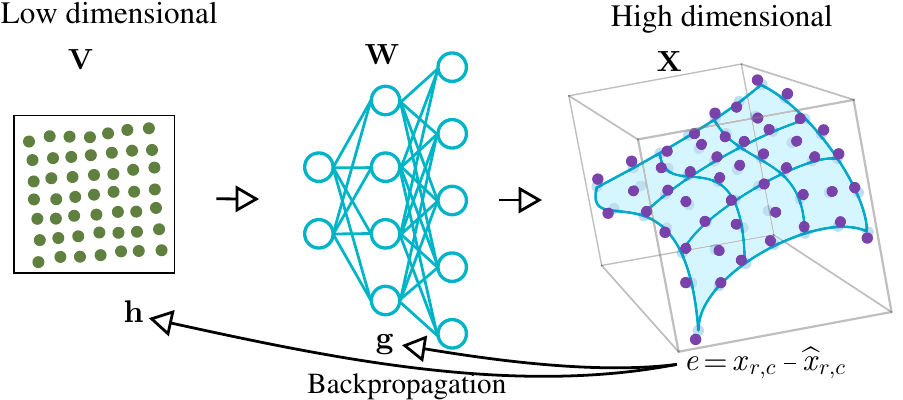}
		\caption{UBP trains an MLP to fit to high-dimensional observations, $\mathbf{X}$. For each known $\mathbf{x}_{r,c} \in \mathbf{X}$, UBP uses backpropagation to compute the gradient vectors $\mathbf{g}$ and $\mathbf{h}$, which are used to update the weights, $\mathbf{W}$, and the input vector $\mathbf{v}_r$.}
		\label{fig_ubp}
	\end{center}
\end{figure*}

In order to formally describe the UBP algorithm, we define the following terms. The relationships between these terms are illustrated graphically in Figure~\ref{fig_ubp}.

\begin{singlespacedlist}
	\item Let $\mathbf{X}$ be a given $n\times d$ matrix, which may have many missing elements. We seek to impute values for these elements. $n$ is the number of instances. $d$ is the number of attributes.
	\item Let $\mathbf{V}$ be a latent $n\times t$ matrix, where $t < d$.
	\item If $x_{r,c}$ is the element at row $r$, column $c$ in $\mathbf{X}$, then $\hat{x}_{r,c}$ is the value predicted by the MLP for this element when $\mathbf{v}_r\in\mathbf{V}$ is fed forward into the MLP.
	\item Let $w_{ij}$ be the weight that feeds from unit $i$ to unit $j$ in the MLP.
	\item For each network unit $i$ on hidden layer $j$, let $\beta_{j,i}$ be the net input into the unit, let $\alpha_{j,i}$ be the output or activation value of the unit, and let $\delta_{j,i}$ be an error term associated with the unit.
	\item Let $l$ be the number of hidden layers in the MLP.
	\item Let $\mathbf{g}$ be a vector representing the gradient with respect to the weights of an MLP, such that $g_{i,j}$ is the component of the gradient that is used to refine $w_{i,j}$.
	\item Let $\mathbf{h}$ be a vector representing the gradient with respect to the inputs of an MLP, such that $h_i$ is the component of the gradient that is used to refine $v_{r,i}\in\mathbf{v}_r$.
\end{singlespacedlist}

Using backpropagation to compute $\mathbf{g}$, the gradient with respect to the weights, is a common operation for training MLPs \citep{rumelhart:backpropagation,werbos:backpropagation}.
Using backpropagation to compute $\mathbf{h}$, the gradient with respect to the inputs, however, is much less common, so we provide a derivation of it here.
In this deriviation, we compute each $h_i\in\mathbf{h}$ from the presentation of a single element $x_{r,c}\in\mathbf{X}$.
It could also be derived from the presentation of a full row (which is typically called ``on-line training"), or from the presentation of all of $\mathbf{X}$ (``batch training"), but since we assume that $\mathbf{X}$ is high-dimensional and is missing many values, it is significantly more efficient to train with the presentation of each known element individually.
We begin by defining an error signal, $E=(x_{r,c}-\hat{x}_{r,c})^2$, and then express the gradient as the partial derivative of this error signal with respect to the inputs: 
\begin{equation}
	h_i=\frac{\partial E}{\partial v_{r,i}}.
	\label{eq_begin}
\end{equation}
The intrinsic input $v_{r,i}$ affects the value of $E$ through the net value of a unit $\beta_i$ and further through the output of a unit $\alpha_i$.
Using the chain rule, Equation \ref{eq_begin} becomes:
\begin{equation}
	h_i=\frac{\partial E}{\partial \alpha_{0,c}}\frac{\partial \alpha_{0,c}}{\partial \beta_{0,c}}\frac{\partial \beta_{0,c}}{\partial v_{r,i}}.
\end{equation}
The backpropagation algorithm calculates $\frac{\partial E}{\partial \alpha_{0,c}}\frac{\partial \alpha_{0,c}}{\partial \beta_{0,c}}$ (which is $\frac{\partial E}{\partial \beta_{j,i}}$ for a network unit) as the error term $\delta_{j,i}$ associated with a network unit.
Thus, to calculate $h_i$ the only additional calculation that needs to be made is $\frac{\partial \beta_j}{\partial v_{r,i}}$.
For a network with 0 hidden layers:
$$\frac{\partial \beta_{0,c}}{\partial v_{r,i}} = \frac{\partial}{\partial v_{r,i}} \sum_t w_{t,c}\: v_{r,t},$$ 
which is non-zero only when $t$ equals $i$ and is equal to $w_{i,c}$.
When there are no hidden layers ($l=0$):
\begin{equation}
	h_i=-w_{i,c} \delta_c.
	\label{eq_no_hidden}
\end{equation}
If there is at least one hidden layer ($l > 0$), then,
$$\frac{\partial \beta_{0,c}}{\partial v_{r,i}} = \frac{\partial \beta_{0,c}}{\partial\alpha_1} \frac{\partial \alpha_1}{\partial \beta_1}\dots\frac{\partial \alpha_l}{\partial \beta_l}\frac{\partial \beta_l}{v_{r,i}},$$
where the $\alpha_k$ and $\beta_k$ represent the output values and the net values for the units in the $k^\text{th}$ hidden layer.
As part of the error term for the units in the $l^\text{th}$ layer, backpropagation calculates $\frac{\partial \beta_{0,c}}{\partial\alpha_1} \frac{\partial \alpha_1}{\partial \beta_1}\dots\frac{\partial \alpha_l}{\partial \beta_l}$ as the error term associated with each network unit.
Thus, the only additional calculation for $h_i$ is:
$$\frac{\partial \beta_{l}}{\partial v_{r,i}} = \frac{\partial}{\partial v_{r,i}} \sum_j \sum_t w_{j,t}\: v_{r,t}.$$
As before, $\frac{\partial \beta_{l}}{\partial v_{r,i}}$ is non-zero only when $t$ equals $i$.
For network with at least one hidden layer:
\begin{equation}
	h_i=-\sum_j w_{i,j} \delta_j.
	\label{eq_with_hidden}
\end{equation}
Equation~\ref{eq_with_hidden} is a strict generalization of Equation~\ref{eq_no_hidden}.
Equation~\ref{eq_no_hidden} only considers the one output unit, $c$, for which a known target value is being presented, whereas
Equation~\ref{eq_with_hidden} sums over each unit, $j$, into which the intrinsic value $v_{r,i}$ feeds.

\subsection{3-phase Training}

UBP trains $\mathbf{V}$ and $\mathbf{W}$ in three phases:
1) the first phase computes an initial estimate for the intrinsic vectors, $\mathbf{V}$,
2) the second phase computes an initial estimate for the network weights, $\mathbf{W}$, and
3) the third phase refines them both together.
All three phases train using stochastic gradient descent, which we derive from the classic backpropagation algorithm.
We now briefly give an intuitive justification for this approach. In our initial experimentation, we used the simpler approach of training in a single phase.
With several problems, we observed that early during training, the intrinsic point vectors, $\mathbf{v}_i\in\mathbf{V}$, tended to separate into clusters.
The points in each cluster appeared to be unrelated, as if they were arbitrarily assigned to one of the clusters by their random initialization.
As training continued, the MLP effectively created a separate mapping for each cluster in the intrinsic representation to the corresponding values in $\mathbf{X}$.
This effectively places an unnecessary burden on the MLP, because it must learn a separate mapping from each cluster that forms in $\mathbf{V}$ to the expected target values.
In phase 1, we give the intrinsic vectors a chance to self-organize while there are no hidden layers to form nonlinear separations among them.
Likewise, phase 2 gives the weights a chance to organize without having to train against moving inputs.
These two preprocessing phases initialize the system (consisting of both intrinsic vectors and weights) to a good initial starting point, such that gradient descent is more likely to find a local optimum of higher quality.
Our empirical results validate this theory by showing that UBP produces more accurate imputation results than NLPCA, which only refines $\mathbf{V}$ and $\mathbf{W}$ together.

\begin{algorithm}[tb]
	\caption{UBP($\mathbf{X}$)}
	\label{alg_ubp}
	\begin{algorithmic}[1]
		\STATE Initialize each element in $\mathbf{V}$ with small random values.
		\STATE Let $\mathbf{T}$ be the weights of a single-layer perceptron
		\STATE Initialize each element in $\mathbf{T}$ with small random values.
		\STATE $\eta' \leftarrow 0.01$; $\eta'' \leftarrow 0.0001$; $\gamma \leftarrow 0.00001$; $\lambda \leftarrow 0.0001$
		\STATE $\eta \leftarrow \eta'$; $s'\leftarrow\infty$
		\WHILE {$\eta > \eta''$}
			\STATE $s\leftarrow$ train\_epoch($\mathbf{X},\mathbf{T},\lambda,\true,0$)
			\STATE {\bf if} $1-s/s'<\gamma$ {\bf then} $\eta \leftarrow \eta / 2$
			\STATE $s'\leftarrow s$
		\ENDWHILE
		\STATE Let $\mathbf{W}$ be the weights of a multi-layer perceptron with $l$ hidden layers, $l \ge 0$
		\STATE Initialize each element in $\mathbf{W}$ with small random values.
		\STATE $\eta \leftarrow \eta'$; $s'\leftarrow\infty$
		\WHILE {$\eta > \eta''$}
			\STATE $s\leftarrow$ train\_epoch($\mathbf{X},\mathbf{W},\lambda,\false,l$)
			\STATE {\bf if} $1-s/s'<\gamma$ {\bf then} $\eta \leftarrow \eta / 2$
			\STATE $s'\leftarrow s$
		\ENDWHILE
		\STATE $\eta \leftarrow \eta'$; $s'\leftarrow\infty$
		\WHILE {$\eta > \eta''$}
			\STATE $s\leftarrow$ train\_epoch($\mathbf{X},\mathbf{W},0,\true,l$)
			\STATE {\bf if} $1-s/s'<\gamma$ {\bf then} $\eta \leftarrow \eta / 2$
			\STATE $s'\leftarrow s$
		\ENDWHILE
		\STATE {\bf return} $\{\mathbf{V},\mathbf{W}\}$
	\end{algorithmic}
\end{algorithm}

Pseudo-code for the UBP algorithm, which trains $\mathbf{V}$ and $\mathbf{W}$ in three phases, is given in Algorithm~\ref{alg_ubp}.
This algorithm calls Algorithm~\ref{alg_refine}, which performs a single epoch of training.
A detailed description of Algorithm~\ref{alg_ubp} follows.

A matrix containing the known data values, $\mathbf{X}$, is passed in to UBP (See Algorithm~\ref{alg_ubp}).
UBP returns $\mathbf{V}$ and $\mathbf{W}$.
$\mathbf{V}$ is a matrix such that each row, $\mathbf{v}_i$, is a low-dimensional representation of the corresponding row, $\mathbf{x}_i$.
$\mathbf{W}$ is a set or ragged matrix containing weight values for an MLP that maps from each $\mathbf{v}_i$ to an approximation of $\mathbf{x}_i\in\mathbf{X}$.
$\mathbf{v}_i$ may be forward-propagated into this MLP to estimate values for any missing elements in $\mathbf{x}_i$.

{\bf Lines~1-9} perform the first phase of training, which computes an initial estimate for $\mathbf{V}$.

{\bf Line~1} of Algorithm~\ref{alg_ubp} initializes the elements in $\mathbf{V}$ with small random values.
Our implementation draws values from a Normal distribution with a mean of 0 and a deviation of 0.01.

{\bf Lines~2-3} initialize the weights, $\mathbf{T}$, of a single-layer perceptron using the same mechanism.
This single-layer perceptron is a temporary model that is only used in phase 1 to assist the initial training of $\mathbf{V}$.

{\bf Line~4} sets some heuristic values that we used to detect convergence.
We note that many other techniques could be used to detect convergence.
Our implementation, used the simple approach of dividing half of the training data for a validation set.
We decay the learning rate whenever predictions fail to improve by a sufficient amount on the validation data.
This simple approch always stops, and it yielded better empirical results than a few other variations that we tried.
$\eta'$ specifies an initial learning rate. Convergence is detected when the learning rate falls below $\eta''$.
$\gamma$ specifies the amount of improvement that is expected after each epoch, or else the learning rate is decayed.
$\lambda$ is a regularization term that is used during the first two phases to ensure that the weights do not become excessively saturated before the final phase of training.
No regularization is used in the final phase of training because we want the MLP to ultimately fit the data as closely as possible.
(Overfit can still be mitigated by limiting the number of hidden units used in the MLP.)
We used the default heuristic values specified on this line in all of our experiments because tuning them seemed to have little impact on the final results.
We believe that these values are well-suited for most problems, but could possibly be tuned if necessary.

{\bf Line~5} sets the learning rate, $\eta$, to the initial value.
The value $s'$ is used to store the previous error score.
As no error has yet been measured, it is initialized to $\infty$.

{\bf Lines~6-9} train $\mathbf{V}$ and $\mathbf{T}$ until convergence is detected.
$\mathbf{T}$ may then be discarded.

{\bf Lines~10-16} perform the second phase of training.
This phase differs from the first phase in two ways: 1) an MLP is used instead of a temporary single-layer perceptron, and 2) $\mathbf{V}$ is held constant during this phase.

{\bf Lines~17-21} perform the third phase of training.
In this phase, the same MLP that is used in phase 2 is used again, but $\mathbf{V}$ and $\mathbf{W}$ are both refined together.
Also, no regularization is used in the third phase.

\subsection{Stochastic gradient descent}

\begin{algorithm}[tb]
	\caption{train\_epoch($\mathbf{X},\mathbf{W},\lambda,p,l$)}
	\label{alg_refine}
	\begin{algorithmic}[1]
		\FOR {{\bf each} known $x_{r,c}\in\mathbf{X}$ in random order}
			\STATE Compute $\alpha_c$ by forward-propagating $\mathbf{v}_r$ into an MLP with weights $\mathbf{W}$.
			\STATE $\delta_c\leftarrow (x_{r,c}-\alpha_c)f'(\beta_c)$
			\FOR {{\bf each} hidden unit $i$ feeding into output unit $c$}
				\STATE $\delta_i\leftarrow w_{i,c} \delta_c f'(\beta_i)$
			\ENDFOR
			\FOR {{\bf each} hidden unit $j$ in an earlier hidden layer (in backward order)}
				\STATE $\delta_j\leftarrow \sum_k w_{j,k} \delta_k f'(\beta_j)$
			\ENDFOR
			\FOR {{\bf each} $w_{i,j}\in\mathbf{W}$}
				\STATE $g_{i,j}\leftarrow -\delta_j \alpha_i$
			\ENDFOR
			\STATE $\mathbf{W} \leftarrow \mathbf{W} - \eta (\mathbf{g} + \lambda\mathbf{W})$
			\IF {$p=\true$}
				\FOR {$i$ {\bf from} $0$ {\bf to} $t-1$}
					\STATE {\bf if} $l=0$ {\bf then} $h_i\leftarrow -w_{i,c} \delta_c$\\ {\bf else} $h_i\leftarrow -\sum_j w_{i,j} \delta_j$
				\ENDFOR
				\STATE $\mathbf{v}_r \leftarrow \mathbf{v}_r - \eta (\mathbf{h} + \lambda\mathbf{v}_r)$
			\ENDIF
		\ENDFOR
		\STATE $s\leftarrow$ measure RMSE with $\mathbf{X}$
		\STATE {\bf return} $s$
	\end{algorithmic}
\end{algorithm}

Next, we describe Algorithm~\ref{alg_refine}, which performs a single epoch of training by stochastic gradient descent.
This algorithm is very similar to an epoch of traditional backpropagation, except that it presents each element individually, instead of presenting each vector, and it conditionally refines the intrinsic vectors, $\mathbf{V}$, as well as the weights, $\mathbf{W}$.

{\bf Line~1} presents each known element $x_{r,c}\in\mathbf{X}$ in random order.

{\bf Line~2} computes a predicted value for the presented element given the current $\mathbf{v}_r$.
Note that efficient implementations of line 2 should only propagate values into output unit $c$.

{\bf Lines~3-7} compute an error term for output unit $c$, and each hidden unit in the network.
The activation of the other output units is not computed, so the error on those units is 0.

{\bf Lines~8-10} refine $\mathbf{W}$ by gradient descent.

{\bf Line 11} specifies that $\mathbf{V}$ should only be refined during phases 1 and 3.

{\bf Lines~12-14} refine $\mathbf{V}$ by gradient descent.

{\bf Line~15} computes the root-mean-squared-error of the MLP for each known element in $\mathbf{X}$.

In order to enable UBP to process nominal (categorical) attributes, we convert such values to a vector representing a membership weight in each category.
For example, a given value of ``cat" from the value set \{``mouse",``cat",``dog"\} is represented with the vector in  $\langle 0, 1, 0 \rangle$.
Unknown values in this attribute are converted to 3 unknown real values, requiring the algorithm to make 3 predictions.
After missing values are imputed, we convert the data back to its original form by finding the mode of each categorical distribution.
For example, the predicted vector $\langle 0.4, 0.25, 0.35 \rangle$ would be converted to a prediction of ``mouse".

\subsection{Complexity}

Because UBP uses a heuristic technique to detect convergence, a full analysis of the computational complexity of UBP is not possible. However,
it is sufficiently similar to other well-known techniques that comparisons can be made. Matrix factorization is generally considered to be a very
efficient imputation technique \citep{koren:matrix_factorization}. Nonlinear PCA is a nonlinear generalization of matrix factorization. If a linear
activation function is used, then it is equivalent to matrix factorization, with the additional (very small) computational overhead of computing the
activation function. When hidden layers are added, computational complexity is increased, but remains proportional to the number of weights in the network.
UBP adds two additional phases of training to Nonlinear PCA. Thus, in the worst case, UBP is 3 times slower than Nonlinear PCA. In practice, however,
the first two phases tend to be very fast (because they optimize fewer values), and these preprocessing phases may even cause the third phase to
be much faster (by initializing the weights and intrinsic vectors to start at a position much closer to a local optimum).

\section{Empirical Validation}
\label{sec_results}



Because running time was not a significant issue with UBP, our empirical validation focuses on imputation accuracy.
We compared UBP with 5 other imputation algorithms.
The imputation methods that we examined as well as their algorithmic parameter values (including UBP) are:
\begin{singlespacedlist}
\item{\bf Mean/Mode or Baseline (BL).} To establish a ``baseline" for comparison, we compare with the method of replacing missing values in continuous attributes with the mean of the non-missing values in that attribute, and replacing missing values in nominal (or categorical) attributes with the most common value in the non-missing values of that attribute.
There are no parameters for BL.
It is expected that any reasonable algorithm should outperform this baseline algorithm with most problems.
\item{\bf Fuzzy K-Means (FKM).} We varied $k$ (the number of clusters) over the set $\{4,8,16\}$, we varied the $L_P$-norm value for computing distance over the set $\{1, 1.5, 2\}$ (Manhattan distance to Euclidean distance), and the fuzzification factor over the set $\{1.3, 1.5\}$ which were reported to be the most effective values \citep{li_fuzzy_k_means}.
\item{\bf Instance-Based Imputation (IBI).} We used cosine correlation to evaluate similarity, and we varied $k$ (the number of neighbors) over the set $\{1, 5, 21\}$.
These values were selected because they were all odd, and spanned the range of intuitively suitable values.
\item{\bf Matrix Factorization (MF).} We varied the number of intrinsic values over the set $\{2,8,16\}$, and the regularization term over the set $\{0.001,0.01,0.1\}$.
Again, these values were selected to span the range of intuitively suitable values.
\item{\bf Nonlinear PCA (NLPCA).} We varied the number of hidden units over the set $\{0,8,16\}$, and the number of intrinsic values over the set $\{2,8,16,32\}$.
In the case of 0 hidden units, only a single layer of sigmoid units was used.
\item{\bf Unuspervised Backpropagation (UBP).} The parameters were varied over the same ranges as those of NLPCA.
\end{singlespacedlist} 
For NLPCA and UBP, we used the logistic function as the activation function and summed squared error as the objective function.
Thus, we imputed missing values in a total of 66000 dataset scenarios.
For each algorithm, we found the set of parameters that yielded the best results, and we compared only these best results for each algorithm averaged over the ten runs of differing random seeds.
To facilitate reproduction of our results, and to assist with related research efforts, we have integrated our implementation of UBP and all of the competitor algorithms into the Waffles machine learning toolkit \citep{gashler2011jmlr}

\begin{figure*}[!tb]
	\begin{center}
		\includegraphics[width=5.5in]{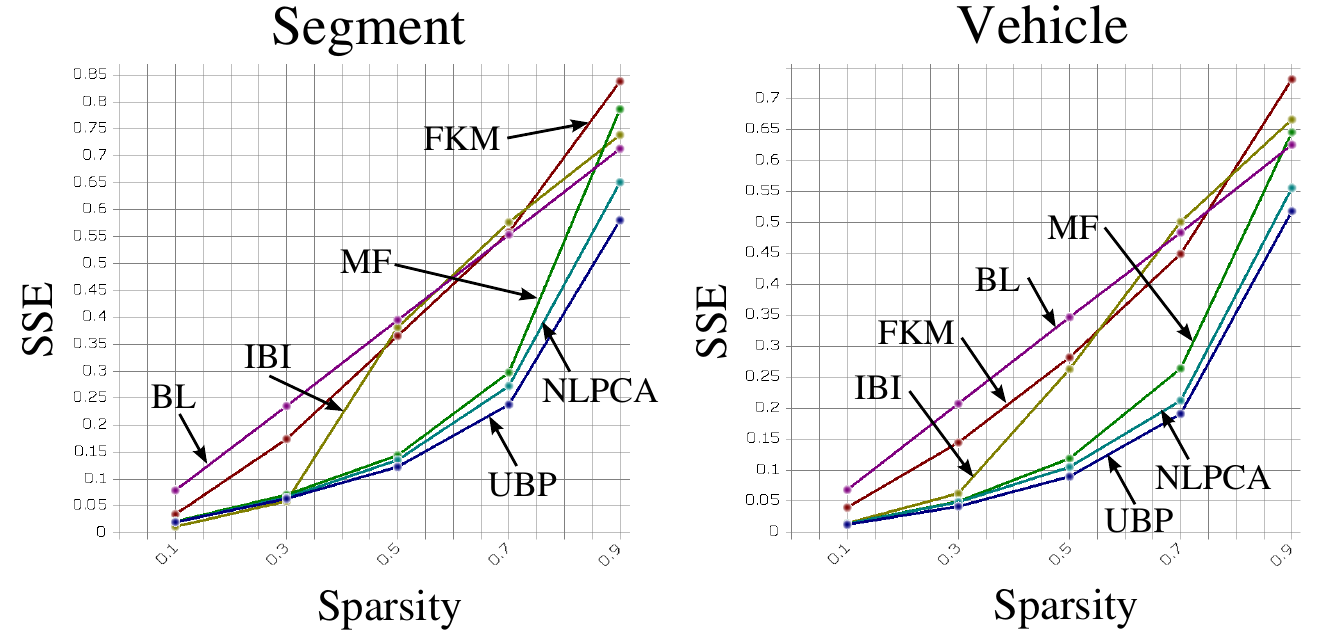}
		\caption{A comparison of the average sum-squared error in each pattern by 5 imputation techniques over a range of sparsity values with two representative datasets. (Lower is better.) }
		\label{fig_charts}
	\end{center}
\end{figure*}

In order to evaluate the effectiveness of UBP and related imputation techniques, we gathered a set of 24 datasets from the UCI repository \citep{uci_repository}, the Promise repository \citep{promise_repository}, and mldata.org: $\{$abalone, arrhythmia, bupa, colic, credit-g, desharnais, diabetes, ecoli, eucalyptus, glass, hypothyroid, ionosphere, iris, nursery, ozone, pasture, sonar, spambase, spectrometer, teaching\_assistant, vote, vowel, waveform-500, and yeast$\}$.
To ensure an objective evaluation, this collection was determined before evaluation was performed, and was not modified to emphasize favorable results. To ensure that our results would be applicable for tasks that require generalization, we removed the class labels from each dataset so that only the input features could be used for imputing missing values.
We normalized all real-valued attributes to fall within a range from 0 to 1 so that every attribute would carry approximately equal weight in our evaluation. We then removed completely at random\footnote{Other categories of ``missingness", besides missing completely at random (MCAR), have been studied \citep{little2002statistical}, but we restrict our analysis to the imputation of MCAR values.} 
$u$\% of the values from each dataset, where $u\in\{10,30,50,70,90\}$.

For each dataset, and for each $u$, we generated 10 datasets with missing values, each using a different random number seed, to make a total of 1200 tasks for evaluation.
The task for each imputation algorithm was to restore these missing values.
We measured error by comparing each predicted (imputed) value with the corresponding original normalized value, summed over all attributes in the dataset.
For nominal (categorical) values, we used Hamming distance, and for real values, we used the squared difference between the original and predicted values.
The average error was computed over all of the patterns in each dataset.

Figure~\ref{fig_charts} shows two representative comparisons of the error scores obtained by each algorithm at varying levels of sparsity.
Comparisons with other datasets generally exhibited similar trends.
MF, NLPCA, and UBP did much better than other algorithms when 50\% or 70\% of the values were missing.
No algorithm was best in every case, but UBP achieved the best score in more cases than any other algorithm.
Table~\ref{table_datasets} summarizes the results of these comparisons.
UBP achieved lower error than the other algorithm in 20 out of 25 pair-wise comparisons, each comparing imputation scores with 24 datasets averaged over 10 runs with different random seeds. In 15 pair-wise comparisons, UBP did better with a sufficient number of datasets to establish statistical significance according to the Wilcoxon Signed Ranks test. These cases are indicated with a ``$\surd$" symbol.

\begin{table}[tbhp]
	\caption{A high-level summary of comparisons between UBP and five other imputation techniques.
		Results are shown for each of the 5 sparsity values. Each row in this table summarizes
		a comparison between UBP and a competitor imputation algorithm for predicting missing values. 
		}
	\label{table_datasets}
	\begin{center}
	\begin{small}
	\begin{sc}
	\begin{tabular}{ccccc}
		\hline
		\inserthalfnewline
		Sparsity & Algorithm	& UBP comparative & P-value\\
		         &             & Wins,Ties,Losses & \\
		\hline
		\inserthalfnewline
\multirow{4}{*}{\begin{sideways}0.1\end{sideways}}	& BL	& {\bf 20},0,4	& 0.001$\surd$\\
			& FKM	& {\bf 19},0,5 	& 0.004$\surd$\\
			& IBI	& {\bf 15},0,9 	& 0.250\ \ \ \ \ \\
			& MF	& 12,0,12 	& 0.348\ \ \ \ \ \\
			& NLPCA & {\bf 11},7,6	& 0.176\ \ \ \ \ \\
		\hline
		\inserthalfnewline
\multirow{4}{*}{\begin{sideways}0.3\end{sideways}}	& BL	& {\bf 20},0,4	& 0.001$\surd$\\
			& FKM	& {\bf 22},0,2	& 0.000$\surd$\\
			& IBI	& {\bf 19},0,5	& 0.022$\surd$\\
			& MF	& 11,0,13	& 0.437\ \ \ \ \ \\
			& NLPCA & {\bf 10},11,3	& 0.036$\surd$\\
		\hline
		\inserthalfnewline
\multirow{4}{*}{\begin{sideways}0.5\end{sideways}}	& BL	& {\bf 20},0,4	& 0.003$\surd$\\
			& FKM	& {\bf 20},0,4	& 0.000$\surd$\\
			& IBI	& {\bf 17},0,7	& 0.018$\surd$\\
			& MF	& 12,0,12	& 0.394\ \ \ \ \ \\
			& NLPCA & {\bf 10},11,3	& 0.046$\surd$\\
		\hline
		\inserthalfnewline
\multirow{4}{*}{\begin{sideways}0.7\end{sideways}}	& BL	& {\bf 16},0,8	& 0.022$\surd$\\
			& FKM	& {\bf 17},0,7	& 0.004$\surd$\\
			& IBI	& {\bf 15},0,9	& 0.040$\surd$\\
			& MF	& {\bf 13},0,11	& 0.312\ \ \ \ \ \\
			& NLPCA & {\bf 7},14,3	& 0.038$\surd$\\
		\hline
		\inserthalfnewline
\multirow{4}{*}{\begin{sideways}0.9\end{sideways}}	& BL	& 7,0,{\bf 17}	& 0.437\ \ \ \ \ \\
			& FKM	& {\bf 17},0,7	& 0.006$\surd$\\
			& IBI	& {\bf 13},0,11	& 0.147\ \ \ \ \ \\
			& MF	& {\bf 16},0,8	& 0.123\ \ \ \ \ \\
			& NLPCA & 2,10,12	& 0.495\ \ \ \ \ \\
		\hline
	\end{tabular}
	\end{sc}
	\end{small}
	\end{center}
\end{table}

To help us further understand how UBP might lead to better classification results, we also performed experiments involving classification with the imputed datasets. We restored class labels to the datasets in which we had imputed feature values. We then used 5 repetitions of 10-fold cross validation with each of 9 learning algorithms from the WEKA \citep{Weka2005} machine learning toolkit: backpropagation (BP), C4.5 \citep{Quinlan1993}, 5-nearest neighbor (IB5), locally weighted learning (LWL) \citep{Atkeson1997}, na\"{i}ve Bayes (NB), nearest neighbor with generalization (NNGE) \citep{Martin1995}, Random Forest (RandF) \citep{breiman:randomforests}, RIpple DOwn Rule Learner (RIDOR) \citep{Gaines1995}, and RIPPER (Repeated Incremental Pruning to Produce Error Reduction) \citep{Cohen1995}.
The learning algorithms were chosen with the intent of being diverse from one another, where diversity is determined using unsupervised meta-learning \citep{Lee2011}.
We evaluated each learning algorithm at each of the previously mentioned sparsity levels using each imputation algorithm with the parameters that resulted in the lowest SSE error score for imputation.
The results from this experiment are summarized for each of the 5 sparsity levels in Tables~\ref{table_class_1} through \ref{table_class_9}.
Each cell in these tables summarizes a set of experiments with an imputation algorithm and classification algorithm pair. The three upper numbers in each cell indicate the number of times that UBP lead to higher, equal, and lower classification accuracy. When the left-most of these three numbers is biggest, this indicates that classification accuracy was higher in more cases when UBP was used as the imputation algorithm.
The lower number in each cell indicates the P-value obtained from the Wilcoxon signed ranks test by performing a pair-wise comparison between UBP and the imputation algorithm of that column. When this value is below 0.05, UBP did better in a sufficient number of pair-wise comparisons to demonstrate that it is better with statistical significance. These cases are indicated with a ``$\surd$" symbol.

\begin{table}[t]
	\caption{Results of classification tests at a sparsity level of 0.1., (meaning
			10\% of data values were removed and then imputed prior to classification). 
			The 3 upper numbers in each cell indicate the wins, ties, and losses for UBP in a pair-wise comparison.
			The lower number in each cell indicates the Wilcoxon signed ranks test P-value evaluated
			over the 24 datasets. (When the value is smaller than 0.05, indicated with a ``$\surd$" symbol, the
			performance of UBP was better with statistical significance.)}
	\label{table_class_1}
	\begin{center}
	\begin{small}
	\begin{sc}
	\begin{tabular}{rccccc}
		\hline
		\inserthalfnewline
		& BL & FKM & IBI & MF & NLPCA\\
		\hline
		\inserthalfnewline
BP     & {\bf 16},1,7  & {\bf 14},0,10 & 12,0,12 & 9,0,{\bf 15}  & 6,7,{\bf 11} \\
       & 0.013$\surd$   & 0.080$\surd$   & 0.835   & 0.874   & 0.915 \\
		\inserthalfnewline
C4.5   & {\bf 16},0,8  & {\bf 18},0,6  & 12,0,12 & {\bf 14},1,9  & {\bf 9},7,8  \\
       & 0.039$\surd$   & 0.011$\surd$   & 0.727   & 0.219   & 0.715 \\
		\inserthalfnewline
IB5    & {\bf 17},1,6  & {\bf 17},0,7  & {\bf 15},0,9  & 11,0,{\bf 13} & {\bf 9},8,7  \\
       & 0.011$\surd$   & 0.005$\surd$   & 0.039$\surd$   & 0.616   & 0.688 \\
		\inserthalfnewline
LWL    & {\bf 16},2,6  & {\bf 12},5,7  & 7,4,{\bf 13}  & 9,2,{\bf 13}  & 7,9,8  \\
       & 0.015$\surd$   & 0.134   & 0.731   & 0.652   & 0.601 \\
		\inserthalfnewline
NB     & 12,0,12 & {\bf 13},0,11 & 12,0,12 & {\bf 13},2,9  & {\bf 9},7,8  \\
       & 0.374   & 0.302   & 0.594   & 0.487   & 0.538 \\
		\inserthalfnewline
Nnge   & {\bf 17},1,6  & {\bf 19},1,4  & {\bf 17},1,6  & {\bf 16},0,8  & {\bf 10},7,7 \\
       & 0.001$\surd$   & 0.000$\surd$   & 0.121   & 0.109   & 0.353 \\
		\inserthalfnewline
\hskip -3mm RandF  & {\bf 17},0,7  & {\bf 16},1,7  & {\bf 14},0,10 & {\bf 16},0,8  & 8,8,8  \\
       & 0.003$\surd$   & 0.006$\surd$   & 0.151   & 0.054   & 0.572 \\
		\inserthalfnewline
Ridor  & {\bf 20},0,4  & {\bf 20},1,3  & 12,0,12 & {\bf 15},0,9  & 6,8,{\bf 10} \\
       & 0.000$\surd$   & 0.000$\surd$   & 0.254   & 0.220   & 0.757 \\
		\inserthalfnewline
\hskip -3mm RIPPER & {\bf 16},1,7  & {\bf 12},1,11 & 12,0,12 & {\bf 15},0,9  & {\bf 10},7,7 \\
       & 0.009$\surd$   & 0.134   & 0.517   & 0.057   & 0.388 \\
		\hline
		\inserthalfnewline
	\end{tabular}
	\end{sc}
	\end{small}
	\end{center}
\end{table}

\begin{table}[tbh]
	\caption{Results of classification tests at a sparsity level of 0.3.} 
	\label{table_class_3}
	\begin{center}
	\begin{small}
	\begin{sc}
	\begin{tabular}{rccccc}
		\hline
		\inserthalfnewline
		& BL & FKM & IBI & MF & NLPCA\\
		\hline
		\inserthalfnewline
BP     & {\bf 18},0,6  & {\bf 18},0,6 & {\bf 15},0,9 & 12,0,12 & 6,10,{\bf 8} \\
       & 0.008$\surd$ & 0.002$\surd$ & 0.048$\surd$ & 0.550 & 0.884 \\
		\inserthalfnewline
C4.5   & {\bf 15},0,9  & {\bf 15},1,8 & {\bf 15},0,9 & {\bf 13},0,11 & 7,10,7 \\
       & 0.024$\surd$ & 0.017$\surd$ & 0.025$\surd$ & 0.461 & 0.735 \\
		\inserthalfnewline
IB5    & {\bf 16},0,8  & {\bf 18},0,6 & 10,0,{\bf 14}& 10,0,{\bf 14} & 4,10,{\bf 10}\\
       & 0.016$\surd$ & 0.004$\surd$ & 0.539 & 0.658 & 0.970 \\
		\inserthalfnewline
LWL    & {\bf 17},1,6  & {\bf 16},2,6 & {\bf 15},2,7 & {\bf 12},2,10 & {\bf 7},12,5 \\
       & 0.005$\surd$ & 0.008$\surd$ & 0.109 & 0.218 & 0.484 \\
		\inserthalfnewline
NB     & 11,0,{\bf 13} & {\bf 13},0,11& 12,0,12& {\bf 13},0,11 & {\bf 9},10,5 \\
       & 0.689 & 0.342 & 0.374 & 0.561 & 0.226 \\
		\inserthalfnewline
Nnge   & {\bf 14},0,10 & {\bf 16},0,8 & {\bf 15},0,9 & {\bf 14},0,10 & 7,10,7 \\
       & 0.132 & 0.020$\surd$ & 0.138 & 0.245 & 0.774 \\
		\inserthalfnewline
\hskip -3mm RandF  & {\bf 14},0,10 & {\bf 17},0,7 & {\bf 15},0,9 & {\bf 13},1,10 & 6,10,{\bf 8} \\
       & 0.132 & 0.021$\surd$ & 0.084 & 0.083 & 0.842 \\
		\inserthalfnewline
Ridor  & {\bf 16},1,7  & {\bf 17},1,6 & {\bf 16},0,8 & {\bf 10},0,14 & 4,10,{\bf 10}\\
       & 0.004$\surd$ & 0.001$\surd$ & 0.018$\surd$ & 0.583 & 0.942 \\
		\inserthalfnewline
\hskip -3mm RIPPER & {\bf 14},0,10 & {\bf 16},0,8 & {\bf 17},0,7 & 9,0,{\bf 15}  & 7,10,7 \\
       & 0.051 & 0.042$\surd$ & 0.026$\surd$ & 0.868 & 0.500 \\
		\hline
		\inserthalfnewline
	\end{tabular}
	\end{sc}
	\end{small}
	\end{center}
\end{table}

\begin{table}[tbh]
	\caption{Results of classification tests at a sparsity level of 0.5.}
	\label{table_class_5}
	\begin{center}
	\begin{small}
	\begin{sc}
	\begin{tabular}{rccccc}
		\hline
		\inserthalfnewline
		& BL & FKM & IBI & MF & NLPCA\\
		\hline
		\inserthalfnewline
BP     & {\bf 13},1,10 & {\bf 14},1,9 & {\bf 20},1,3 & {\bf 14},0,10 & 6,11,{\bf 7} \\
       & 0.055 & 0.033$\surd$ & 0.002$\surd$ & 0.195 & 0.472 \\
		\inserthalfnewline
C4.5   & {\bf 14},0,10 & {\bf 14},0,10& {\bf 14},0,10& {\bf 15},0,9  & {\bf 9},11,4 \\
       & 0.245 & 0.076 & 0.080 & 0.115 & 0.104 \\
		\inserthalfnewline
IB5    & {\bf 15},0,9  & {\bf 17},0,7 & {\bf 16},1,7 & {\bf 13},1,10 & {\bf 8},11,5 \\
       & 0.015$\surd$ & 0.001$\surd$ & 0.002$\surd$ & 0.093 & 0.338 \\
		\inserthalfnewline
LWL    & {\bf 14},1,9  & {\bf 14},2,8 & {\bf 13},2,9 & {\bf 14},1,9  & 4,13,{\bf 7} \\
       & 0.016$\surd$ & 0.009$\surd$ & 0.077 & 0.193 & 0.825 \\
		\inserthalfnewline
NB     & {\bf 13},0,11 & {\bf 13},0,11& {\bf 12},1,11& {\bf 15},0,9  & {\bf 8},11,5 \\
       & 0.302 & 0.322 & 0.458 & 0.363 & 0.132 \\
		\inserthalfnewline
Nnge   & {\bf 13},0,11 & {\bf 15},0,9 & {\bf 18},0,6 & {\bf 15},0,9  & {\bf 7},11,6 \\
       & 0.245 & 0.048$\surd$ & 0.021$\surd$ & 0.084 & 0.338 \\
		\inserthalfnewline
\hskip -3mm RandF  & {\bf 14},0,10 & {\bf 17},0,7 & {\bf 19},0,5 & {\bf 14},1,9  & 7,10,7 \\
       & 0.109 & 0.015$\surd$ & 0.002$\surd$ & 0.202 & 0.401 \\
		\inserthalfnewline
Ridor  & {\bf 15},1,8  & {\bf 16},0,8 & {\bf 19},0,5 & {\bf 13},0,11 & 6,11,{\bf 7} \\
       & 0.022$\surd$ & 0.005$\surd$ & 0.000$\surd$ & 0.395 & 0.758 \\
		\inserthalfnewline
\hskip -3mm RIPPER & {\bf 15},0,9  & {\bf 16},0,8 & {\bf 19},0,5 & {\bf 14},0,10 & {\bf 7},11,6 \\
       & 0.030$\surd$ & 0.028$\surd$ & 0.001$\surd$ & 0.384 & 0.338 \\
		\hline
		\inserthalfnewline
	\end{tabular}
	\end{sc}
	\end{small}
	\end{center}
\end{table}

\begin{table}[tbh]
	\caption{Results of classification tests at a sparsity level of 0.7.}
	\label{table_class_7}
	\begin{center}
	\begin{small}
	\begin{sc}
	\begin{tabular}{rccccc}
		\hline
		\inserthalfnewline
		& BL & FKM & IBI & MF & NLPCA\\
		\hline
		\inserthalfnewline
BP     & {\bf 13},0,11 & {\bf 17},0,7 & {\bf 17},1,6 & {\bf 16},0,8  & {\bf 8},14,2 \\
       & 0.264 & 0.008$\surd$ & 0.001$\surd$ & 0.051 & 0.026$\surd$ \\
		\inserthalfnewline
C4.5   & 9,0,{\bf 15}  & 9,2,{\bf 13} & {\bf 16},0,8 & {\bf 13},0,11 & 5,14,5 \\
       & 0.763 & 0.474 & 0.064$\surd$ & 0.332 & 0.658 \\
		\inserthalfnewline
IB5    & {\bf 14},0,10 & {\bf 17},0,7 & {\bf 15},0,9 & {\bf 17},0,7  & {\bf 8},14,2 \\
       & 0.165 & 0.005$\surd$ & 0.032$\surd$ & 0.008$\surd$ & 0.042$\surd$ \\
		\inserthalfnewline
LWL    & {\bf 14},1,9  & {\bf 16},1,7 & {\bf 18},2,4 & {\bf 14},1,9  & {\bf 6},16,2 \\
       & 0.013$\surd$ & 0.002$\surd$ & 0.000$\surd$ & 0.121 & 0.147 \\
		\inserthalfnewline
NB     & {\bf 16},0,8  & {\bf 15},0,9 & {\bf 14},0,10& 12,0,12 & {\bf 6},14,4 \\
       & 0.104 & 0.203 & 0.539 & 0.506 & 0.305 \\
		\inserthalfnewline
Nnge   & {\bf 12},1,11 & {\bf 15},0,9 & {\bf 14},0,10& {\bf 14},0,10 & {\bf 8},14,2 \\
       & 0.169 & 0.011$\surd$ & 0.068$\surd$ & 0.018$\surd$ & 0.010$\surd$ \\
		\inserthalfnewline
\hskip -3mm RandF  & {\bf 16},0,8  & {\bf 18},0,6 & {\bf 18},0,6 & {\bf 19},0,5  & {\bf 10},12,2 \\
       & 0.017$\surd$ & 0.001$\surd$ & 0.001$\surd$ & 0.002$\surd$ & 0.003$\surd$ \\
		\inserthalfnewline
Ridor  & {\bf 13},1,10 & {\bf 16},0,8 & {\bf 19},0,5 & {\bf 16},0,8  & 4,15,{\bf 5} \\
       & 0.052 & 0.017$\surd$ & 0.000$\surd$ & 0.042$\surd$ & 0.453 \\
		\inserthalfnewline
\hskip -3mm RIPPER & {\bf 13},0,11 & {\bf 15},0,9 & {\bf 18},0,6 & {\bf 14},0,10 & {\bf 8},15,1 \\
       & 0.211 & 0.060 & 0.015$\surd$ & 0.195 & 0.038$\surd$ \\
		\hline
		\inserthalfnewline
	\end{tabular}
	\end{sc}
	\end{small}
	\end{center}
\end{table}

\begin{table}[tbh]
	\caption{Results of classification tests at a sparsity level of 0.9.}
	\label{table_class_9}
	\begin{center}
	\begin{small}
	\begin{sc}
	\begin{tabular}{rccccc}
		\hline
		\inserthalfnewline
		& BL & FKM & IBI & MF & NLPCA\\
		\hline
		\inserthalfnewline
BP     & 12,0,12 & {\bf 14},0,10 & {\bf 16},3,5  & {\bf 16},0,8 & {\bf 8},10,6 \\
       & 0.658 & 0.180 & 0.033$\surd$ & 0.057 & 0.265 \\
		\inserthalfnewline
C4.5   & 8,0,{\bf 16}  & 10,0,{\bf 14} & {\bf 13},0,11 & {\bf 15},0,9 & 6,10,{\bf 8} \\
       & 0.936 & 0.689 & 0.282 & 0.151 & 0.827 \\
		\inserthalfnewline
IB5    & 10,0,{\bf 14} & 11,0,{\bf 13} & 11,0,{\bf 13} & {\bf 17},0,7 & 5,11,{\bf 8} \\
       & 0.880 & 0.637 & 0.658 & 0.032$\surd$ & 0.712 \\
		\inserthalfnewline
LWL    & {\bf 12},1,11 & {\bf 14},1,9  & {\bf 15},1,8  & {\bf 14},1,9 & 4,10,{\bf 10} \\
       & 0.458 & 0.070 & 0.140 & 0.109 & 0.949 \\
		\inserthalfnewline
NB     & {\bf 21},0,3  & {\bf 18},0,6  & {\bf 14},0,10 & {\bf 13},0,11& {\bf 10},10,4 \\
       & 0.000$\surd$ & 0.001$\surd$ & 0.172 & 0.228 & 0.012$\surd$ \\
		\inserthalfnewline
Nnge   & {\bf 14},0,10 & {\bf 17},0,7  & {\bf 14},0,10 & {\bf 18},0,6 & {\bf 8},10,6 \\
       & 0.180 & 0.004$\surd$ & 0.010$\surd$ & 0.008$\surd$ & 0.265 \\
		\inserthalfnewline
\hskip -3mm RandF  & 12,0,12 & {\bf 15},0,9  & {\bf 17},0,7  & {\bf 20},0,4 & {\bf 10},10,4 \\
       & 0.439 & 0.145 & 0.008$\surd$ & 0.001$\surd$ & 0.074 \\
		\inserthalfnewline
Ridor  & 7,0,{\bf 17}  & {\bf 14},0,10 & {\bf 15},0,9  & {\bf 20},0,4 & 7,10,7 \\
       & 0.880 & 0.187 & 0.084 & 0.002$\surd$ & 0.599 \\
		\inserthalfnewline
\hskip -3mm RIPPER & 10,1,{\bf 13} & 12,0,12 & {\bf 18},0,6  & {\bf 19},0,5 & 7,10,7 \\
       & 0.811 & 0.363 & 0.004$\surd$ & 0.010$\surd$ & 0.450 \\
		\hline
		\inserthalfnewline
	\end{tabular}
	\end{sc}
	\end{small}
	\end{center}
\end{table}

Some classification algorithms were more responsive to the improved imputation accuracy that UBP offered than others. For example, UBP appears to have a more beneficial effect on
classification results when used with Random Forest than with na\"{i}ve Bayes. Overall, UBP was demonstrated to be more beneficial as an imputation algorithm than any of the other
imputation algorithms. As expected, NLPCA was the closest competitor. UBP did better than NLPCA in a larger number of cases, but we were not able to demonstrate that it was better
with statistical significance until sparsity level 0.7. It appears from these results that the improvements of UBP over NLPCA generally have a bigger impact when there are more
missing values in the data. This is significant because the difficulty of effective imputation increases with the sparsity of the data.

%
%
%

\begin{figure}[!tb]
	\begin{center}
		\begin{tabular}{c|c}
			 \includegraphics[width=2.6in]{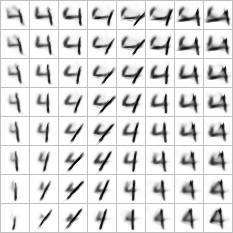} & \includegraphics[width=2.6in]{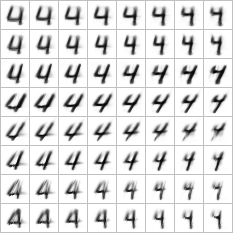}\\
			 a & b\\
		\end{tabular}
		\caption{An MLP trained with NLPCA (a) and UBP(b) on digits from the MNIST dataset of handwritten digits. After training, the digits shown here were
				generated by uniformly sampling over the space of intrinsic values, and generating an image at each sampled intrinsic point.Although both NLPCA and UBP were able
				to generate high-quality digits, UBP was able to organize the various styles more effectively. Note that the height of the horizontal bar varies from left-to-right.
				The results from NLPCA do not provide the ability to vary the height of this bar in each of the possible styles of the digit.}
	\end{center}
	\label{fig_mnist}
\end{figure}

To demonstrate the difference between UBP and NLPCA (and, hence, the advantage of 3-phase training), we briefly examine UBP as a manifold learning (or non-linear dimensionality reduction) technique.
We trained both NLPCA and UBP using data from the MNIST dataset of handwritten digits. We used a multilayer perceptron with a $4\rightarrow8\rightarrow256\rightarrow1$
topology. 2 of the 4 inputs were treated as latent values, and the other 2 inputs we used to specify $(x,y)$ coordinates in each image. The one output was trained to predict
the grayscale pixel value at the specified coordinates. We trained a separate network for each of the 10 digits in the dataset. After training these multilayer perceptrons in this
manner, we uniformly sampled the two-dimensional latent inputs in order to visualize how each multilayer perceptron organized the digits that it had learned. Matrix factorization was not
suitable for this application because it is linear, so it would predict only a linear gradient instead of an image. The other imputation algorithms are not suitable for this task because
they are not designed to be used in a generative manner. Figure~\ref{fig_mnist} shows a sample of '4's generated by NLPCA and UBP.
(Note that the images shown are not part of the original training data. Rather, each was generated by the multilayer perceptron after it was trained. Because the multilayer perceptron
is a continuous function, these digits could be generated with arbitrary resolution, independent of the original training data.)

Both algorithms did approximately equally well at generating digits that appear natural over a range of styles. The primary difference between these results is how they ultimately
formed order in the intrinsic values that represent the various styles. In the case of UBP, somewhat better intrinsic organization is formed. Three distinct styles can be observed in
three horizontal stripes in Figure~\ref{fig_mnist}b: boxey digits in the top two rows, slanted digits in the middle rows, and digits with a closed top in the bottom two rows. The
height of the horizontal bar in the digit varies clearly from left-to-right. Along the left, the horizontal bar crosses at a low point, and along the right, the horizontal bar crosses
at a high point. In the case of NLPCA, similar styles were organized into clusters, but they do not exhibit the same degree of organization that is apparent with UBP. For example, it
does not appear to be able to vary the height of the horizontal bar in each of the three main styles of this digit. This occurs because NLPCA does not have an extra step
designed explicitly to promote organization in the intrinsic values.

This demonstration also serves to show that techniques for imputation and techniques for manifold learning are merging. In the case of these handwritten digits, the multilayer perceptron
was flexible enough to generate visibly appealing digits, even when the intrinsic organization was poor. However, better generalization can be expected when the mapping is simplest, which
occurs when the intrinsic values are better organized. Hence, as imputation and manifold learning are applied to increasingly complex problems, the importance of finding good organization
in the intrinsic values will increase. Future work in this area will explore other techniques for promoting organization within the intrinsic values, and contrast them with the simple
approach proposed by UBP.

%
				to generate high-quality digits, UBP was able to organize the various styles more effectively. Note that the height of the horizontal bar varies from left-to-right.
				The results from NLPCA do not provide the ability to vary the height of this bar in each of the possible styles of the digit.

\section{Conclusions}
\label{sec_conclusions}

Since the NetFlix competition, matrix factorization and related linear techniques have generally been considered to be the \emph{state-of-the-art} in imputation.
Unfortunately, the focus on imputing with a few large datasets has caused the community to largely overlook the value of nonlinear imputation techniques.
One of the contributions of this paper is the observation that Nonlinear PCA, which was presented almost a decade ago, consistently outperforms matrix factorization
across a diversity of datasets. Perhaps this has not yet been recognized because a thorough comparison between the two techniques across a diversity of datasets was not previously performed.

The primary contribution of this paper, however, is an improvement to Nonlinear PCA, which improves its accuracy even more.
This contribution consists of a 3-phase training process that initializes both weights and intrinsic vectors to better starting positions.
We theorize that this leads to better results because it bypasses many local optima in which the network could otherwise settle, and places it closer to the global optimum.

We empirically compared results from UBP with 5 other imputation techniques, including baseline, fuzzy $k$-means, instance-based imputation, matrix factorization, and Nonlinear PCA, with 24 datasets across a range of parameters for each algorithm.
UBP predicts missing values with lower error than any of these other methods in the majority of cases.
We also demonstrated that using UBP to impute missing values leads to better classification accuracy than any of the other imputation techniques over all, especially at higher levels of sparsity, which is the most important case for imputation.
We demonstrated that UBP is also better-suited for manifold learning than NLPCA with a problem involving handwritten digits. Ongoing research seeks to demonstrate that better
organization in the intrinsic variables naturally leads to better generalization. This has significant potential application in unsupervised learning tasks such as automation.


\bibliographystyle{coin}%
\bibliography{refs}

\begin{thebibliography}{}
\providecommand{\natexlab}[1]{#1}
\providecommand{\url}[1]{{#1}}
\expandafter\ifx\csname urlstyle\endcsname\relax
  \providecommand{\doi}[1]{#1}\else
  \providecommand{\doi}{\begingroup \urlstyle{rm}\Url}\fi

\bibitem[\protect\citeauthoryear{Acu{\~n}a and Rodriguez}{Acu{\~n}a and
  Rodriguez}{2004}]{Acuna2004}
{\sc Acu{\~n}a, Edgar}, and {\sc Caroline Rodriguez}. 2004.
\newblock The treatment of missing values and its effect on classifier
  accuracy.
\newblock {\em In\/}~Classification, Clustering, and Data Mining Applications.
  Springer Berlin / Heidelberg, pp.\  639--647.

\bibitem[\protect\citeauthoryear{Adomavicius and Tuzhilin}{Adomavicius and
  Tuzhilin}{2005}]{adomavicius:recommender_systems}
{\sc Adomavicius, G.}, and {\sc A.~Tuzhilin}. 2005.
\newblock {Toward the next generation of recommender systems: A survey of the
  state-of-the-art and possible extensions}.
\newblock IEEE transactions on knowledge and data engineering,~{\bf
  17\/}(6):734--749.
\newblock ISSN 1041-4347.

\bibitem[\protect\citeauthoryear{{Alireza Farhangfar}}{{Alireza
  Farhangfar}}{2008}]{Farhangfar2008}
{\sc {Alireza Farhangfar}, Lukasz Kurgan Jennifer~Dy}. 2008.
\newblock Impact of imputation of missing values on classification error for
  discrete data.
\newblock Pattern Recognition,~{\bf 41}:3692--3705.

\bibitem[\protect\citeauthoryear{Atkeson, Moore, and Schaal}{Atkeson
  et~al.}{1997}]{Atkeson1997}
{\sc Atkeson, Christopher~G.}, {\sc Andrew~W. Moore}, and {\sc Stefan Schaal}.
  1997.
\newblock Locally weighted learning.
\newblock Artificial Intelligence Review,~{\bf 11}:11--73.
\newblock ISSN 0269-2821.

\bibitem[\protect\citeauthoryear{Bengio, Lamblin, Popovici, and
  Larochelle}{Bengio et~al.}{2007}]{Bengio2007}
{\sc Bengio, Yoshua}, {\sc Pascal Lamblin}, {\sc Dan Popovici}, and {\sc Hugo
  Larochelle}. 2007.
\newblock Greedy layer-wise training of deep networks.
\newblock {\em In\/}~NIPS,  pp.\  153--160.

\bibitem[\protect\citeauthoryear{Bengio, Schwenk, Sen{\'e}cal, Morin, and
  Gauvain}{Bengio et~al.}{2006}]{bengio:neural_language_generative_backprop}
{\sc Bengio, Y.}, {\sc H.~Schwenk}, {\sc J.S. Sen{\'e}cal}, {\sc F.~Morin}, and
  {\sc J.L. Gauvain}. 2006.
\newblock Neural probabilistic language models.
\newblock {\em In\/}~Innovations in Machine Learning. Springer, pp.\  137--186.

\bibitem[\protect\citeauthoryear{Breiman}{Breiman}{2001}]{breiman:randomforest%
s}
{\sc Breiman, L.} 2001.
\newblock Random forests.
\newblock Machine Learning,~{\bf 45\/}(1):5--32.

\bibitem[\protect\citeauthoryear{Coheh and Shawe-Taylor}{Coheh and
  Shawe-Taylor}{1990}]{coheh:label_images_generative_backprop}
{\sc Coheh, D.}, and {\sc J.~Shawe-Taylor}. 1990.
\newblock Daugman's gabor transform as a simple generative back propagation
  network.
\newblock Electronics Letters,~{\bf 26\/}(16):1241--1243.

\bibitem[\protect\citeauthoryear{Cohen}{Cohen}{1995}]{Cohen1995}
{\sc Cohen, William~W.} 1995.
\newblock Fast effective rule induction.
\newblock {\em In\/}~In Proceedings of the Twelfth International Conference on
  Machine Learning,  Morgan Kaufmann, pp.\  115--123.

\bibitem[\protect\citeauthoryear{Erhan, Manzagol, Bengio, Bengio, and
  Vincent}{Erhan et~al.}{2009}]{Erhan2009}
{\sc Erhan, Dumitru}, {\sc Pierre-Antoine Manzagol}, {\sc Yoshua Bengio}, {\sc
  Samy Bengio}, and {\sc Pascal Vincent}. 2009.
\newblock The difficulty of training deep architectures and the effect of
  unsupervised pre-training.
\newblock Journal of Machine Learning Research - Proceedings Track,~{\bf
  5}:153--160.

\bibitem[\protect\citeauthoryear{Frank and Asuncion}{Frank and
  Asuncion}{2010}]{uci_repository}
{\sc Frank, A.}, and {\sc A.~Asuncion}. 2010.
\newblock {UCI} machine learning repository.
\newblock \url{http://archive.ics.uci.edu/ml}.

\bibitem[\protect\citeauthoryear{Gaines and Compton}{Gaines and
  Compton}{1995}]{Gaines1995}
{\sc Gaines, Brian~R.}, and {\sc Paul Compton}. 1995.
\newblock Induction of ripple-down rules applied to modeling large databases.
\newblock Journal of Intelligent Information Systems,~{\bf 5}:211--228.
\newblock ISSN 0925-9902.

\bibitem[\protect\citeauthoryear{Gashler, Ventura, and Martinez}{Gashler
  et~al.}{2008}]{gashler:manifoldsculpting}
{\sc Gashler, Michael}, {\sc Dan Ventura}, and {\sc Tony Martinez}. 2008.
\newblock Iterative non-linear dimensionality reduction with manifold
  sculpting.
\newblock {\em In\/}~Advances in Neural Information Processing Systems 20. MIT
  Press, pp.\  513--520.

\bibitem[\protect\citeauthoryear{Gashler}{Gashler}{2011}]{gashler2011jmlr}
{\sc Gashler, Michael~S.} 2011.
\newblock Waffles: A machine learning toolkit.
\newblock Journal of Machine Learning Research,~{\bf MLOSS 12}:2383--2387.

\bibitem[\protect\citeauthoryear{Hinton}{Hinton}{1988}]{hinton:generative_back%
propagation}
{\sc Hinton, G.~E.} 1988.
\newblock Generative back-propagation.
\newblock {\em In\/}~Abstracts 1st INNS.

\bibitem[\protect\citeauthoryear{Hinton and Salakhutdinov}{Hinton and
  Salakhutdinov}{2006}]{autoencoders}
{\sc Hinton, G.~E.}, and {\sc R.~R. Salakhutdinov}. 2006.
\newblock Reducing the dimensionality of data with neural networks.
\newblock Science,~{\bf 28\/}(313 (5786)):504--507.

\bibitem[\protect\citeauthoryear{Jones}{Jones}{1996}]{Jones1996}
{\sc Jones, Michael~P.} 1996.
\newblock Indicator and stratificaion methods for missing explanatory variables
  in multiple linear regression.
\newblock Journal of the American Statistical Association,~{\bf 91}:222--230.

\bibitem[\protect\citeauthoryear{Koren, Bell, and Volinsky}{Koren
  et~al.}{2009}]{koren:matrix_factorization}
{\sc Koren, Y.}, {\sc R.~Bell}, and {\sc C.~Volinsky}. 2009.
\newblock {Matrix factorization techniques for recommender systems}.
\newblock Computer,~{\bf 42\/}(8):30--37.
\newblock ISSN 0018-9162.

\bibitem[\protect\citeauthoryear{Lakshminarayan, Harp, Goldman, Samad,
  et~al.}{Lakshminarayan et~al.}{1996}]{lakshminarayan1996imputation}
{\sc Lakshminarayan, K.}, {\sc S.A. Harp}, {\sc R.~Goldman}, {\sc T.~Samad},
  and {\sc others}. 1996.
\newblock Imputation of missing data using machine learning techniques.
\newblock {\em In\/}~Proceedings of the Second International Conference on
  Knowledge Discovery and Data Mining,  pp.\  140--145.

\bibitem[\protect\citeauthoryear{Lee and Giraud-Carrier}{Lee and
  Giraud-Carrier}{2011}]{Lee2011}
{\sc Lee, Jun}, and {\sc Christophe Giraud-Carrier}. 2011.
\newblock A metric for unsupervised metalearning.
\newblock Intelligent Data Analysis,~{\bf 15\/}(6):827--841.

\bibitem[\protect\citeauthoryear{Li, Deogun, Spaulding, and Shuart}{Li
  et~al.}{2004}]{li_fuzzy_k_means}
{\sc Li, Dan}, {\sc Jitender Deogun}, {\sc William Spaulding}, and {\sc Bill
  Shuart}. 2004.
\newblock Towards missing data imputation: A study of fuzzy k-means clustering
  method.
\newblock {\em In\/}~Rough Sets and Current Trends in Computing, Volume 3066 of
  {\em Lecture Notes in Computer Science}. Springer Berlin / Heidelberg, pp.\
  573--579.

\bibitem[\protect\citeauthoryear{Li, Chen, and Lin}{Li
  et~al.}{2008}]{li:filtering_techniques}
{\sc Li, Q.}, {\sc Y.P. Chen}, and {\sc Z.~Lin}. 2008.
\newblock {Filtering Techniques For Selection of Contents and Products}.
\newblock {\em In\/}~Personalization of Interactive Multimedia Services: A
  Research and development Perspective,  Nova Science Publishers.
\newblock ISBN 978-1-60456-680-2.

\bibitem[\protect\citeauthoryear{Little and Rubin}{Little and
  Rubin}{2002}]{little2002statistical}
{\sc Little, R.J.A.}, and {\sc D.B. Rubin}. 2002.
\newblock {Statistical analysis with missing data}.
\newblock Wiley series in probability and mathematical statistics. Probability
  and mathematical statistics. Wiley.
\newblock ISBN 9780471183860.

\bibitem[\protect\citeauthoryear{Luengo}{Luengo}{2011}]{luengo:thesis2011}
{\sc Luengo, J.} 2011.
\newblock Soft Computing based learning and Data Analysis: Missing Values and
  Data Complexity.
\newblock Ph.\ D. thesis, Department of Computer Science and Artificial
  Intelligence, University of Granada.

\bibitem[\protect\citeauthoryear{Martin}{Martin}{1995}]{Martin1995}
{\sc Martin, Brent}. 1995.
\newblock Instance-based learning: nearest neighbour with generalisation.
\newblock Technical Report 95/18, University of Waikato, Department of Computer
  Science.

\bibitem[\protect\citeauthoryear{Quinlan}{Quinlan}{1989}]{quinlain_missing_val%
ues}
{\sc Quinlan, J.~R.} 1989.
\newblock Unknown attribute values in induction.
\newblock {\em In\/}~Proceedings of the sixth international workshop on Machine
  learning,  pp.\  164--168.

\bibitem[\protect\citeauthoryear{Quinlan}{Quinlan}{1993}]{Quinlan1993}
{\sc Quinlan, J.~Ross}. 1993.
\newblock C4.5: Programs for Machine Learning.
\newblock Morgan Kaufmann, San Mateo, CA, USA.

\bibitem[\protect\citeauthoryear{Rubin}{Rubin}{1987}]{rubin:1987}
{\sc Rubin, D.~B.} 1987.
\newblock Multiple Imputation for Nonresponse in Surveys.
\newblock Wiley.

\bibitem[\protect\citeauthoryear{Rumelhart, Hinton, and Williams}{Rumelhart
  et~al.}{1986}]{rumelhart:backpropagation}
{\sc Rumelhart, D.E.}, {\sc G.E. Hinton}, and {\sc R.J. Williams}. 1986.
\newblock {Learning representations by back-propagating errors}.
\newblock Nature,~{\bf 323}:9.

\bibitem[\protect\citeauthoryear{Sarwar, Karypis, Konstan, and Reidl}{Sarwar
  et~al.}{2001}]{sarwar:item_based_collab_filter}
{\sc Sarwar, B.}, {\sc G.~Karypis}, {\sc J.~Konstan}, and {\sc J.~Reidl}. 2001.
\newblock {Item-based collaborative filtering recommendation algorithms}.
\newblock {\em In\/}~Proceedings of the 10th international conference on World
  Wide Web,  ACM.
\newblock ISBN 1581133480. pp.\  285--295.

\bibitem[\protect\citeauthoryear{Sayyad~Shirabad and Menzies}{Sayyad~Shirabad
  and Menzies}{2005}]{promise_repository}
{\sc Sayyad~Shirabad, J.}, and {\sc T.J. Menzies}. 2005.
\newblock {The {PROMISE} Repository of Software Engineering Databases.}
\newblock School of Information Technology and Engineering, University of
  Ottawa, Canada.
\newblock \url{http://promise.site.uottawa.ca/SERepository}.

\bibitem[\protect\citeauthoryear{Schafer and Graham}{Schafer and
  Graham}{2002}]{schafer2002missing}
{\sc Schafer, J.L.}, and {\sc J.W. Graham}. 2002.
\newblock {Missing data: Our view of the state of the art.}
\newblock Psychological methods,~{\bf 7\/}(2):147.
\newblock ISSN 1939-1463.

\bibitem[\protect\citeauthoryear{Scholz, Kaplan, Guy, Kopka, and Selbig}{Scholz
  et~al.}{2005}]{nlpca}
{\sc Scholz, M.}, {\sc F.~Kaplan}, {\sc C.~L. Guy}, {\sc J.~Kopka}, and {\sc
  J.~Selbig}. 2005.
\newblock Non-linear pca: a missing data approach.
\newblock Bioinformatics,~{\bf 21\/}(20):3887--3895.

\bibitem[\protect\citeauthoryear{Shafer}{Shafer}{1997}]{Shafer1997}
{\sc Shafer, J.~L.} 1997.
\newblock Analysis of Incomplete Multivariate Data.
\newblock Chapman and Hall, London.

\bibitem[\protect\citeauthoryear{Tak{\'a}cs, Pil{\'a}szy, N{\'e}meth, and
  Tikk}{Tak{\'a}cs et~al.}{2009}]{takacs2009scalable}
{\sc Tak{\'a}cs, G.}, {\sc I.~Pil{\'a}szy}, {\sc B.~N{\'e}meth}, and {\sc
  D.~Tikk}. 2009.
\newblock Scalable collaborative filtering approaches for large recommender
  systems.
\newblock The Journal of Machine Learning Research,~{\bf 10}:623--656.

\bibitem[\protect\citeauthoryear{Tenenbaum, de~Silva, and Langford}{Tenenbaum
  et~al.}{2000}]{tenenbaum:isomap}
{\sc Tenenbaum, Joshua~B.}, {\sc Vin de~Silva}, and {\sc John~C. Langford}.
  2000.
\newblock A global geometric framework for nonlinear dimensionality reduction.
\newblock Science,~{\bf 290}:2319--2323.

\bibitem[\protect\citeauthoryear{Werbos}{Werbos}{1990}]{werbos:backpropagation}
{\sc Werbos, P.J.} 1990.
\newblock {Backpropagation through time: What it does and how to do it}.
\newblock Proceedings of the IEEE,~{\bf 78\/}(10):1550--1560.
\newblock ISSN 0018-9219.

\bibitem[\protect\citeauthoryear{Witten and Frank}{Witten and
  Frank}{2005}]{Weka2005}
{\sc Witten, Ian~H.}, and {\sc Eibe Frank}. 2005.
\newblock {Data Mining: Practical machine learning tools and techniques} (2nd
  ed.).
\newblock Morgan Kaufmann, San Fransisco.

\bibitem[\protect\citeauthoryear{Zhang and Wang}{Zhang and
  Wang}{2007}]{zhang:mlle}
{\sc Zhang, Z.}, and {\sc J.~Wang}. 2007.
\newblock {MLLE: Modified locally linear embedding using multiple weights}.
\newblock Advances in Neural Information Processing Systems,~{\bf 19}:1593.
\newblock ISSN 1049-5258.

\end{thebibliography}

\label{lastpage}

\end{document}